# Features or Shape? Tackling the False Dichotomy of Time Series Classification[†]

Sara Alaee[*]    Alireza Abdoli[*]    Christian Shelton[*]    Amy C. Murillo[¶]    Alec C. Gerry[¶]

Eamonn Keogh[*]

**Abstract**

Time series classification is an important task in its own right, and it is often a precursor to further downstream analytics. To date, virtually all works in the literature have used *either* shape-based classification using a distance measure or feature-based classification after finding some suitable features for the domain. It seems to be underappreciated that in many datasets it is the case that some classes are best discriminated with features, while others are best discriminated with shape. Thus, making the shape vs. feature choice will condemn us to poor results, at least for some classes. In this work, we propose a new model for classifying time series that allows the use of both shape and feature-based measures, when warranted. Our algorithm automatically decides which approach is best for which class, and at query time chooses which classifier to trust the most. We evaluate our idea on real world datasets and demonstrate that our ideas produce statistically significant improvement in classification accuracy.

## 1 Introduction

The problem of time series classification has been an active research area for decades [2][3][5][7][9]. There are two classic approaches for time series classification in the literature: shape-based classification and feature-based classification. Shape-based classification determines the best class according to a distance measure (e.g. Euclidean Distance) between the unlabeled exemplar and a set prototypes to represent the classes [1][3][23]. Feature-based classification, on the other hand, finds the best class according to the set of features defined for the time series [5][13][14][27]. These features measure properties of the time series (e.g. standard deviation, mean, complexity, fractal dimension etc.). Then the unlabeled exemplar time series is classified based on its nearest neighbor in the feature space or using a decision tree or other eager learner.

Given the existence of the two rival philosophies, which one should we use? It is possible that:

1) One of the two techniques dominates for all problems. However, a quick inspection of the literature [7], or a little introspection convinces us otherwise. There are clearly problems for which one of the two approaches is much better on all classes.
2) On a problem-by-problem basis, one of the techniques dominates. For example, perhaps for electrocardiograms *shape* is the best, but for electroencephalograms *feature* is the best. This seems to be the implicit assumption of most of the community [13].

However, there is a third possibility that seems to have escaped the attention of the community:

3) On a single problem, it might be possible that one of the techniques is better for one subset of the classes, and the other technique is better for another subset. For example, perhaps for ECGs [25][26], *shape* is better at distinguishing between the two classes of `PVCs` and `VTs`, but shape cannot distinguish between `PVCs` and `VF`. Whereas, some feature *can* tease `PVCs` and `VF` apart.

Given this third possibility, it is clear that neither of the two techniques will dominate for some problems, but that some "combination" of both might be the best. Do such domains really exist? In fact, we believe that they are very common. For example, in Figure 1 we show an example of such problem.

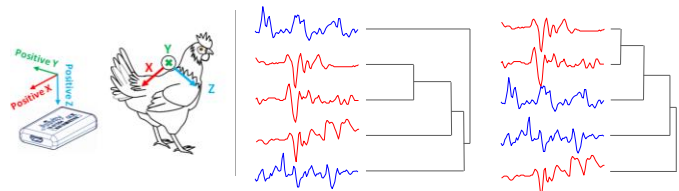

Figure 1: Scientists collect telemetry from domestic chickens, to count the frequency of `dustbathing` and `pecking`. Three examples of chicken `pecking` behavior (*Red*) clustered with two `dustbathing` behaviors (*Blue*), using (*left*) "Euclidean Distance" and (*right*) the "Complexity" feature.

As we will later show, this task of classifying chicken behavior from accelerometer data is essentially impossible to solve with a single modality of classification. The *peck* behavior has highly conserved shape. While the `dustbathing` behavior does not have a stereotypical shape, it *is* possible to recognize by several features, including "complexity" [15][23][28]. Moreover, we believe that many problems have this property. To consider a more familiar example, for human behavior, `walking` can be differentiated from `running` or `walking-upstairs` based on shape, but we surely cannot expect to separate `sitting` from `reclining` based on shape.

In this work, we introduce a framework that can learn to choose the best classification approach for any given dataset on a class-by-class basis. If a dataset is shape "friendly" (meaning that shape alone can classify the time series), then the framework will utilize only shape. If it is feature "friendly", then the framework degenerates to feature based classification. However, if a combination of shape and feature works best for a dataset on a class-by-class basis, then the framework uses an appropriate combination of both approaches.

We will demonstrate the utility of our ideas on real-world problems and show that we can significantly out-perform rival methods that consider only one modality.

[*]Computer Science and Engineering Department, University of California, Riverside, USA.
[¶]Department of Entomology, University of California, Riverside, USA.
{salae001, aabdo002, cshelton, amy.murillo, alecg, eamonn}@ucr.edu
[†]This work is supported by NSF IIS-1510741, 1631776, CPS 1544969 and gifts from MERL Labs, NetAPP, and a Google faculty award.

The rest of this paper is organized as follows. In Section 2 we present the motivation. Section 3 describes the formal definitions and background. In Section 4 we outline our approach. Section 0 contains an extensive experimental evaluation. Finally, we offer conclusions and thoughts on future directions in Section 6.

## 2 Motivation

Because our underlying claim is novel, we will take the time to show an explicit numerical example. Our example is semi-synthetic for clarity, but as we will later show, highly illustrative of real-world datasets.

Suppose we have a classification problem with four classes of time series, namely `Gun`, `NoGun`, `Random-noise` and `Random-walk` as shown in Figure 2. The `Gun` and `NoGun` classes are from the classic Gun dataset of the UCR time series data mining archive [8], with 100 instances, each of length 150 data points. We randomly selected 20 instances of each class. We generated the same number of instances with the same length for both `Random-noise` and `Random-walk`.

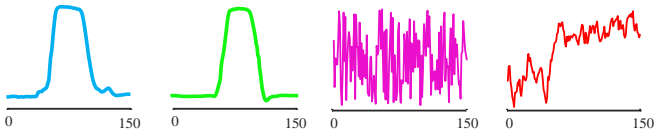

Figure 2: *Blue*) A `Gun` instance where an actor is aiming a toy gun. *Green*) A `NoGun` instance where an actor is pointing a finger at a target. *Pink*) `Random-noise`. *Red*) `Random-walk`.

Using the one-leave-out cross validation method, we first classified these instances using one-nearest-neighbor Euclidean distance, showing the results in Table 1.

Table 1: Classification Results using Shape (Euclidean Distance)

|  |  | PREDICTED | | | |
|---|---|---|---|---|---|
|  |  | Gun | NoGun | Random-noise | Random-walk |
| ACTUAL | Gun | 19 | 1 | 0 | 0 |
|  | NonGun | 1 | 19 | 0 | 0 |
|  | Random-noise | 0 | 0 | 0 | 20 |
|  | Random-walk | 1 | 0 | 0 | 19 |

Table 1 shows that classification using Z-normalized Euclidean distance (i.e. shape-based classification) works well for the {`Gun` , `NoGun`} subset of this problem. However, every instance of the `Random-noise` class is misclassified as `Random-walk`. In retrospect, this is not surprising. The curse of dimensionality tells us that a random vector tends to be far from all other objects [4]. Indeed, because random vectors tend to "live" on the edge of the space, and high dimensional spaces are almost all "edge", the distance between any two random vectors tends to be close to the diameter of the space. However, the random walks are relatively smooth, and reside towards the center of the space. Thus, the nearest neighbor for any instance in `Random-noise` will tend to be an instance from the `Random-walk` class.

In contrast, consider Table 2. Here we attempt classification using one-nearest-neighbor with the *complexity* feature [23]. This feature can trivially distinguish between the highly complex `Random-noise` class and the relatively simple `Random-walk` class. However, as we might expect, there is little reason to expect that complexity can differentiate between `Gun` and `NoGun` since these two classes are visually very similar.

Table 2: Classification Results using Feature (Complexity)

|  |  | PREDICTED | | | |
|---|---|---|---|---|---|
|  |  | Gun | NoGun | Random-noise | Random-walk |
| ACTUAL | Gun | 14 | 6 | 0 | 0 |
|  | NonGun | 8 | 12 | 0 | 0 |
|  | Random-noise | 0 | 0 | 20 | 0 |
|  | Random-walk | 0 | 0 | 0 | 20 |

The practical upshot of this is that shape-based classification achieves a 29% error-rate and feature-based classification achieves a 17% error rate. Suppose we had an oracle meta-algorithm, which computed both the shape-based and feature-based classification, but "magically" chose the prediction of the correct algorithm. This oracle algorithm could achieve only 2.5% error rate.

This experiment hints at our idea. If we can have a "semi-oracle" algorithm that dynamically chooses the best classification approach at query time, and can choose better than random guessing, then we could do better than either of the single modality approaches.

**2.1 The Limit and Scope of our Claims.** Before moving on, it is worth the time to make our claim more concrete and explain why it may have escaped the community's attention.

For shape-based classification there are many distance measures we can use to measure shape similarity, including Euclidean Distance, Dynamic Time Warping etc. [1][7][16]. In addition, while shape-based measures are often used with a nearest neighbor classifier, they can be used with decision trees, Support Vector Machines or even as inputs to various deep learning algorithms. Similar remarks apply to feature-based learning, where there are a plethora of features and algorithms proposed [13]. We make no claims about which combination is best for any given problem. We believe that our observations have the potential to improve any current choice possible. Concretely, our claim is limited to the following:

**Claim**: There exist datasets for which an approach that is expressive enough to consider *both* shape and feature, can outperform the better of using *just* shape or feature.

Our claim is orthogonal to which distance measure, features, or classifier is used. A practical implication of our claim is that it automatically defines the appropriate strawman to compare to. In every case, the correct comparison to our combined Shape-Feature algorithm, is the *identical* algorithm, but with just shape, and with just feature.

Finally, it is interesting to ask why this apparently low-hanging fruit has not been exploited before, especially given the very active research community in time series classification. While this is speculation, we believe that an overreliance on the UCR archive may be partly to blame. The vast majority of papers on time series classification consider some improvements on some of the UCR datasets as a necessary and sufficient condition for publication [1][16]. However, as [22] recently noted, many of the datasets in the archive were created by using a shape-based measure to extract exemplars from a longer time series. Thus, it is tautological to find that shape-based measures work well on this ubiquitous benchmark. To bypass this issue, and to truly stress test our approach, in this work we will attempt classification on continuous raw unedited real-world datasets.

## 3 Definitions and Background

We begin by defining the key terms used in this work. The data we work with is a *time series*.

DEFINITION 3.1. A *time series* $T$ is a sequence of real-valued numbers $t_i$: $T = [t_1, t_2, \ldots, t_n]$ where $n$ is the length of $T$.

Typically, neither shape nor feature based classifiers consider the entire time series to make a prediction, but instead consider only local *subsequences* of the times series.

DEFINITION 3.2. A *subsequence* $T_{i,m}$ of a time series $T$ is a continuous subset of data points from $T$ of length $m$ starting at position $i$. $T_{i,m} = [t_i, t_{i+1}, \ldots, t_{i+m-1}], 1 \leq i \leq n-m+1$.

The length of the subsequence is typically set by the user based on domain knowledge. For example, for most human actions, ½ second may be appropriate, but for classifying transient stars, three days may be appropriate.

Our basic plan is to classify the time series using a combination of feature and shape measures. It is generally accepted that the *nearest neighbor classifier* is one of the best options for time series data [1].

DEFINITION 3.3. The *nearest neighbor (nn) classifier* for a time series $T$ is an algorithm that for each query $Q$ finds its nearest neighbor in $T$, and assigns $Q$ that neighbor's class.

In case of a shape-based classifier, the nearest neighbor is defined using either the Euclidean distance (ED) or the Dynamic Time Warping (DTW) as the distance measure. The distance between a query (i.e. an unlabeled exemplar) and all the other subsequences in the time series is stored in a vector called the *distance profile*.

DEFINITION 3.4. A *distance profile* $D$ is a vector of distances between a given query $Q$ and each subsequence in the time series.

Figure 3 illustrates calculating the distance profile (D). The distance measure is the Euclidean distance between Z-normalized subsequences [18][20]. The value of the distance profile is low wherever the subsequence is highly similar to the query. In case the query Q is part of the time series, the value for the distance profile at the location of query is zero, and close to zero just before and after. To avoid such trivial matches an exclusion zone with the length of $\frac{m}{2}$ ($m$ is the length of the query) is assumed to the left and right of the location of the query [18]. The distance profile can be computed very efficiently using the MASS algorithm [6].

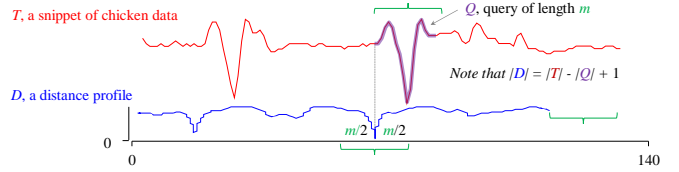

Figure 3: Distance profile $D$ obtained by searching for query $Q$ in time series $T$. Here the query is extracted from the time series itself.

While distance is the measure of similarity between two subsequences in a shape-based classifier, the feature-based classifier finds similarity based on a set of *features*.

DEFINITION 3.5. Given $T$ as the time series and $X$ as a function, the *feature vector F* reports the value of $X$ for every subsequence of $T$. Each feature vector corresponds to a measurable property of the time series.

## 4 Our Proposed Approach

Our task-at-hand reduces to the following steps:

1. *Identify the set of useful features*, allowing the possibility that different classes may best be distinguished with different subsets of features (including the "feature" of shape).
2. *Build the K independent models*, one per class.
3. *Set the K relevant thresholds* (given the user's class-dependent tolerance for false positives/false negatives)
4. *Deploy the model*

This algorithm outlined in **Table 5** has two subroutines outlined in **Table 3**. Individual elements are motivated and explained in the following subsections.

**4.1 Identify the Features.** Instead of a *global* model that looks at all the features, we propose a set of K individual *local* models. These local binary models can be considered experts in recognizing a particular class. Because of their limited task, they can choose to *only* consider the features (including the 'feature' of *shape*) that best recognizes that class. The choice of which features to use is beyond the scope of this paper. Fulcher et. al. [13] presented a library of over 9000 features that can be used to quantify time series properties including the classic features such as min, max, standard deviation, etc. For simplicity, we limit the features to those that have been previously introduced in the data mining literature [14][24].

**4.2 Build the Independent Models.** We are given a time series, the set of classes describing the time series and the set of features (including the feature of *shape*) describing each individual class. The classification process is outlined in Table 5. For each class, the algorithm calls the subroutine outlined in Table 3 to compute distributions (line 2).

Table 3: Algorithm for computing distribution

```
   Input:  D, Train data
   Input:  F, Feature list
   Input:  L, Labels
   Input:  C, class
   Input:  excZone, length of exclusion zone
   Output: M_p, M_n, distributions of class and non-class
1  for i ← 1:numberOfFeatures
2    [PF,ind] ← sort(generateProfile(D,F[i]))
3    for j ← 1: numberOfSnippets
4      if C exists in L[ind[j]:ind[j]+ excZone]
5        P_list ← [P_list; PF[ind[j]]]
6        L[ind[j]:ind[j]+ excZone] ← 'None'
7      else
8        N_list ← [N_list; PF[ind[j]]]
9      end
10   end
11   M_p = [M_p; Histogram(P_list)]
12   M_n = [M_n; Histogram(N_list)]
13 end
```

For every feature in the set, the subroutine computes a feature vector for the data (line 2). The feature value corresponding to each subsequence that includes an instance of the given class is saved in a list (P_list), while all the other values are saved in another list (N_list) (lines 3-10). We assumed an exclusion zone between each two subsequences that contain an instance. The exclusion zone is an empty region where we expect no class exists. For example, if we used a smartwatch's accelerometer to detect when a user takes a drag from a cigarette, we should set an exclusion zone of seven minutes, otherwise another detection surely represents the *same* cigarette. These exclusion window lengths can be set by common sense (especially for human activities), learned from strongly labeled data, or set using domain knowledge. For example, for the chicken dataset we describe later, the literature suggests that two examples of pecking should be at least ¼ of a second apart [15][21].

Lines 11 and 12 calculate and return the distributions of values in P_list and N_list.

Given these distributions, the classification algorithm calculates the probability of each class for the test data using the subroutine outlined in Table 4. For each subsequence with a certain feature value, the subroutine computes and returns the probability based on the frequency of that feature value in the distribution vectors (lines 1-15). This probability is called a *local* model.

Lines 7-9 (Table 5) combine the local models using the Naïve Bayes formula (4.1) to generate the combined model.

$$(4.1) \quad P(C_i|\{f_1(t), f_2(t), \ldots, f_n(t)\}) = \frac{P(C_i|f_1(t))P(C_i|f_2(t))\ldots P(C_i|f_n(t))}{P(C_i)}$$

Where $f_j(t)$ is the function calculating feature $j$ on the subsequence starting at $t$ and $P(C_i|f_{j\leq n}(t))$ is the local model for class $C_i$.

Table 4: Algorithm for computing probability

```
   Input:  Hist_P, distribution of feature values for class
   Input:  Hist_N, distribution of feature values for non-class
   Output: p, the probability distribution of the given feature for the given class
1  for j ← 1: numberOfSnippets
2    targetBin_p ← bin_containing_the_snippet_in_Hist_P
3    targetBin_n ← bin_containing_the_snippet_in_Hist_N
4    if targetBin_p is not empty
5      prob_p ← density(targetBin_p)
6    else
7      prob ← [prob;'small_value']
8    end
9    if targetBin_n is not empty
10     prob_n ← density(targetBin_n)
11   else
12     prob ← [prob;'small_value']
13   end
14   prob ← [prob; prob_p/(prob_p+prob_n)]
15 end
```

**4.3 Set the Relevant Thresholds.** For each class, we need a threshold that best defines our relative tolerance for the false positives vs. false negatives. The thresholds can be either manually adjusted by the user (static) or learned through a feedback loop (dynamic), assuming ground truth labels are eventually available. In the dynamic case, the user may inspect the results produced by multiple runs of the algorithm and choose the threshold setting corresponding to the most desired point on the ROC curve.

Table 5: Algorithm for classifying the time series

```
   Input:  D_train, train data; D_test, test data
   Input:  Q, Vector of queries for different classes
   Input:  L, Labeled time series
   Input:  F, Feature list
   Input:  C, List of classes
   Input:  excZone, exclusion zones for different classes
   Input:  THR, vector of thresholds for different classes
   Output: pl, the predicted labels of the time series
1  for i ← 1:numberOfClasses
2    [M_p,M_n] ← compDist(D_train,F,Q[i],L,C[i],excZone[i]) //
3    See Table 3
4    for i ← 1:numberOfFeatures
5      PF ← generateProfile(D_test,F[i])
6      prob ← compProb(M_p[i],M_n[i],PF) // See Table 4
7      Pr ← Pr.*prob
8    end
9    p ← [p;Pr]
10 end
11 prob ← p.* THR
12 for j ← 1:numberOfSnippets
13   [max_prob, max_class] ← max(prob(j))
14   pl (j) ← max_class
15   pl (j+1:j+ excZone[max_class]) ← 'None'
16 end
```

**4.4 Deploy the Model.** Line 11 applies the threshold on the combined model to generate the final overarching model for each class. Formally:

$$(4.2) \quad P(C_i|f_1(t), f_2(t), \ldots, f_n(t)).Thrshld_i$$

The algorithm classifies the data based on this final model (lines 12-16). Subsequence starting at $t$ is classified as $C_i$ if its final probability is higher for $C_i$ than any other classes.

## 5 Empirical Evaluation

To ensure that our experiments are reproducible, we have created a supporting website which contains all data including the codes and results [12].

**5.1 Performance Evaluation.** An overview of the two datasets is presented in Table 6. The datasets used are the real datasets which include real behaviors. The human dataset is somewhat contrived, in that the individual performed proscribed behaviors in a given order [10]. Naturally the chicken dataset is not contrived in any way.

In principle, a single behavior could have two or more possible instantiations which are semantically identical. For example, a one-pump and a three-pump handshake would both be in class handshake but have very different shapes. For simplicity, in this work we assume that there is a single way to perform a behavior. However, generalizing the code to a polymorphic dictionary is trivial.

Likewise, for simplicity, in both datasets we limit our attention to binary classification problems. However, our algorithm does not make any assumptions about the number of classes. With appropriate set of features, the model can be easily generalized to the multiclass case. In a sense, these are not true binary problems, because each dataset has an implicit Other class. In the case of the Chicken dataset, the Other class is the majority of the dataset.

Table 6: Dataset Characteristics

| Dataset | Source | Classes # | Train length | Test length |
|---------|--------|-----------|--------------|-------------|
| PAMAP | [10] | 2 | 14,000 | 14,000 |
| Chicken | [15] | 2 | 8,658,539 | Short: 176,154<br>Long: 8,637,971 |

Another important point is that our datasets, as we elaborate in the following sections, are both weakly labeled, meaning that every annotated region contains one or more of the specified class. In addition, there are almost certainly instances of a class outside the annotated regions which the annotator failed to label. Moreover, we do not know the exact number of instances of a class inside a region. To address this issue, we utilize the concept of Multiple Instance Learning (MIL) [17]. MIL assumes each annotated region as a "bag" containing one or more instances of a class. If at least a single instance of a class is matched inside a bag, it is treated as a true positive. If no instances of the class are detected inside the bag, then the entire bag is treated as a false negative. In case an instance of class is mismatched inside a bag belonging to some other behavior, then it is treated as a false positive. Finally, if no mismatch occurs inside a bag of a non-relevant class, then the entire bag is treated as a true negative.

**5.2 Case Study: Physical Activity Dataset.** The PAMAP project (Physical Activity Monitoring for Aging People) monitors the physical activity of elderly people [10]. The dataset is comprised of long multidimensional time series, with each dimension showing a signal from an accelerometer located on a certain part of the body. Figure 4 shows examples of some of these activities.

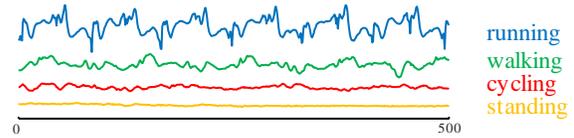

Figure 4: Examples of some types of outdoor activities from x-accelerometer located on the chest

Each activity is *weakly* labeled, meaning that every annotated region may contain more than one behavior. For example, an instance of running may contain regions of walking and standing, as the athlete pauses briefly at an intersection or watering station etc.

For simplicity, we considered data from only one sensor (the X-axis accelerometer on the chest). We selected three types of activities (i.e. standing, walking and running). We picked several instances of each activity from the same subject and concatenated them together to make a long time series. Figure 5 shows the train data (of length 14,000 data points) and Figure 7(a) shows the test data (of length 14,000 data points), both created in the same way. Regions of running have been interpolated between activities only as transitions (or other class).

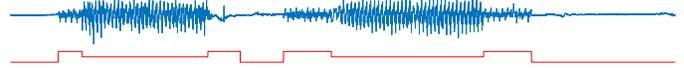

Figure 5: PAMAP train data (*Blue*) with annotations (*Red*). Each bar represents one activity region. The bars with the same heights represent the same types of activities.

Using the algorithm described in Section 4, we first built the models. Figure 6 shows a visual summary of the models.

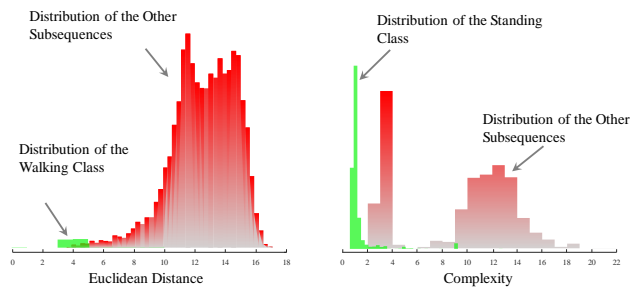

Figure 6: Walking is distinguished from other subsequences with the Euclidean distance (*left*), while standing is distinguished from other subsequences using the complexity feature (*right*).

As the figure shows, each of the walking and standing activities have a dominant model with which they can almost be distinguished from other activities. Shape-based model is dominant for walking, because walking has a well-preserved shape in the entire data set. In contrast, feature-based model can better describe standing. We could describe this class with several features such as low variance, low amplitue, non-periodic, etc. Since "complexity" combines most of these attributes into one feature, we used it in our model. Using this model, we classified the test data. Figure 7(b) shows a visual summary of the results. We can see that the combined model works well for this dataset.

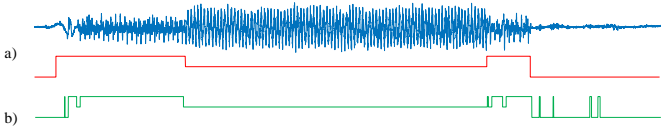

Figure 7: PAMAP test data (*Blue*) with annotations (*Red*). The bars with the same heights represent the same types of activities. Despite confusions between activities in some regions, our algorithm (*Green*) correctly found almost all instances of `walking` and `standing`.

Table 7 and Table 8 present the confusion matrices for `walking` and `standing` classes.

Table 7: Confusion Matrix for Walking

|  |  | PREDICTED | |
|---|---|---|---|
|  |  | **Walking** | **Non-Walking** |
| **ACTUAL** | **Walking** | 2 | 0 |
|  | **Non-Walking** | 1 | 2 |

The results in Table 7 indicate that our classification algorithm achieves 66% precision and 100% recall in matching instances of `walking`. Overall, the classifier has 80% accuracy for the `walking` activity.

It is important to restate that we do not measure accuracy by the number of correct and incorrect instances of a class. Instead, we count the number of bags. More precisely, if at least one instance of a class was correctly classified inside the bag corresponding to that class, we consider that bag as a true positive. The opposite is the case for false positives. As Figure 7(b) shows, our algorithm worked visually well in classifying the instances of `walking`, meaning that it found almost all true positives.

Table 8: Confusion Matrix for Standing

|  |  | PREDICTED | |
|---|---|---|---|
|  |  | **Standing** | **Non-Standing** |
| **ACTUAL** | **Standing** | 2 | 0 |
|  | **Non-Standing** | 0 | 3 |

The results in Table 8 indicate that our algorithm achieves 100% recall and 100% precision for `Standing`. The overall accuracy is 100% for the `Standing` activity.

As mentioned earlier, the correct way to compare our combined Shape/Feature algorithm, is to compare its results to the same algorithm, but with *just* shape, and with *just* feature. Let us see what the results would be if we use only a shape-based classifier and only a feature-based classifier for this dataset. Figure 8 shows the results.

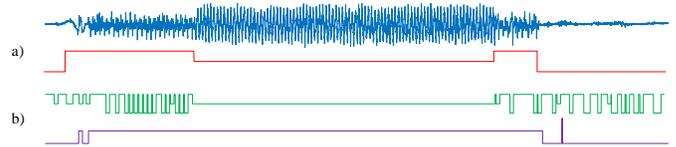

Figure 8: PAMAP test data (*Blue*) with annotations (*Red*). The shape-only version (*Green*) found most `walking` instances but it missed many instances of `standing`. The feature-only version (*Purple*) found almost all `standings`. However, it couldn't classify instances of `walking`.

As shown in Figure 8(b), the shape-only version found almost all instances of `walking`. However, it missed most instances of `standing`. On the other hand, with the feature-only version, the algorithm correctly classified almost all instances of `standing`. However, it could not distinguish `walking` from `running` (the non-class). Table 9 compares the performance of all three models. As shown in the table, the shape-based algorithm works as well for `walking` as the combined algorithm. However, it is much worse for `standing`. In contrast, feature-based algorithm works as well for `standing` as the combined algorithm, while it is worse for `walking`. The reason why accuracy is high for `walking` with the feature is because the algorithm did not confuse `walking` with `standing`. However, it confused almost all instances of `walking` with `running` (false positives and false negatives), due to similar complexity profiles. Thus, the precision and recall are both zero.

Table 9: Performance summary of different models for PAMAP

|  | **ACTIVITY** | **PRECISION** | **RECALL** | **ACCURACY** |
|---|---|---|---|---|
| **SHAPE** | Walking | 0.5 | 1 | 0.6 |
|  | Standing | 0.33 | 0.5 | 0.4 |
| **FEATURE** | Walking | 0 | 0 | 0.4 |
|  | Standing | 1 | 1 | 1 |
| **COMBINED** | Walking | 0.66 | 1 | 0.8 |
|  | Standing | 1 | 1 | 1 |

**5.3 Case Study: Chicken Dataset.** We investigated the utility of our algorithm for separating distinct chicken behaviors. A healthy chicken is expected to display a set of behaviors [11][19][21]. In this work, we only considered two of them, i.e. `pecking` and `dustbathing`. `Pecking` is the act of striking the ground with beak, while `dustbathing` is the act of grooming by rolling in the dirt. We used the dataset introduced into the community by [15].

The training data as illustrated in Figure 9 is a 24-hour one-dimensional time series of length 8,658,539 datapoints (*Blue*) with its corresponding annotations (*Red*). Each label corresponds to one behavior and there are regions of non-behaviors between every two behaviors.

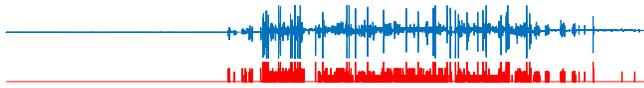

Figure 9: Twenty-four hours chicken train data (*Blue*) with the annotations (*Red*). Each bar represents one behavior zone. The bars with the same heights represent the same behaviors.

Figure 10 shows a visual summary of the models created for the train data. Here again, one class (i.e. `pecking`) has a dominant shape-based model, while the other one (i.e. `dustbathing`) has a dominant feature-based model.

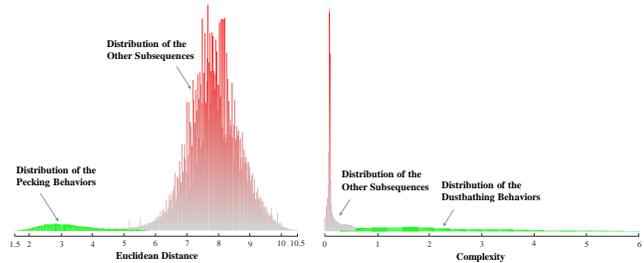

Figure 10: `Pecking` is distinguished from other subsequences with the Euclidean distance (left), while `dustbathing` is distinguished from other subsequences using the complexity feature (right).

We ran our classification algorithm against two versions of the test data: a short version and a long version. The short version, illustrated in Figure 11(a) is a one-dimensional time series of length 30 minutes, inspected and labeled manually by a veterinary entomologist expert. The labels correspond to the *regions* where one or more instances of the same behaviors occurred. Figure 11(b) shows a visual summary of the classification results.

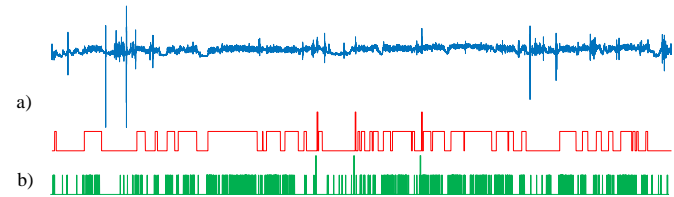

Figure 11: Thirty minutes chicken test data (*Blue*) with annotations (*Red*). The bars with the same heights represent the same types of activities. Our algorithm (*Green*) correctly found all regions of `dustbathing` and most instances of `pecking`.

Table 10 provides the confusion matrix for the performance of `pecking`. Given the results in this table, our classification model has a 57% precision and 100% recall in matching instances of the `pecking` behavior. Overall, the classifier has 63% accuracy for the `pecking`.

We asked an expert to further review our results. This is what he noted about false positives: "*I inspected the dataset for false positive `pecking` behaviors. I reviewed 90 objects manually and only 15 of them looked false positives while the rest (75) looked like good pecks which have escaped human labeling.*" The 15 mentioned false positives were located in only 12 bags out of 76. With this expert annotation, the updated precision, recall and accuracy of `pecking` are 75%, 100% and 84%, respectively.

Table 10: Confusion Matrix for Pecking Behavior

|  |  | PREDICTED | |
|---|---|---|---|
|  |  | Pecking | Non-Pecking |
| ACTUAL | Pecking | 37 | 0 |
|  | Non-Pecking | 28 | 11 |

Table 11 provides the confusion matrix for the performance of `dustbathing`. This time, our classification model achieved a 100% precision and 100% recall in matching instances of the `dustbathing` behavior. Overall, the classifier has 100% accuracy for the `dustbathing`, which is a favorable result.

Table 11: Confusion Matrix for Dustbathing Behavior

|  |  | PREDICTED | |
|---|---|---|---|
|  |  | Dustbathing | Non-Dustbathing |
| ACTUAL | Dustbathing | 3 | 0 |
|  | Non-Dustbathing | 0 | 73 |

To demonstrate the superiority of the combined model over either of the shape and the feature models, let us see what the results would be if we use only a shape-based classifier or a feature-based classifier for this dataset. Figure 12 shows the results.

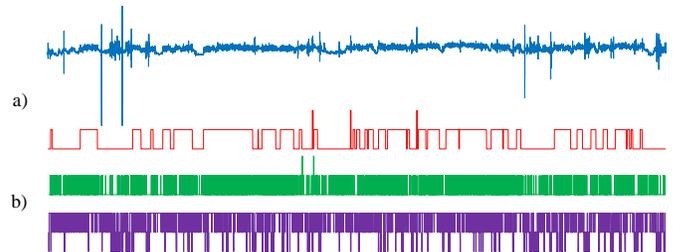

Figure 12: Thirty minutes chicken test data (*Blue*) with annotations (*Red*). The shape-only version of our algorithm (*Green*) found all instances of `pecking` but it missed almost all instances of `dustbathing`. The feature-only version (*Purple*) found all instances of `dustbathing` but it missed many instances of `pecking`.

As is evident from Figure 12(b), the shape-only scenario found all instances of `pecking`, but it missed almost all instances of `dustbathing` (except one). In contrast, the feature-only scenario found all instances of `dustbathing` while missing many instances of `pecking`. Table 12 provides a brief performance summary of all three models.

Table 12: Performance summary of different models for Chickens

|  | BEHAVIOR | PRECISION | RECALL | ACCURACY |
|---|---|---|---|---|
| SHAPE | Pecking | 0.49 | 1 | 0.5 |
|  | Dustbathing | 0.5 | 0.3 | 0.96 |
| FEATURE | Pecking | 0.3 | 0.43 | 0.25 |
|  | Dustbathing | 0.05 | 1 | 0.25 |
| COMBINED | Pecking | 0.75 | 1 | 0.84 |
|  | Dustbathing | 1 | 1 | 1 |

Even though recall is high for `pecking` in shape-based version and for `dustbathing` in feature-based version, the combined model seems to beat the other two in terms of precision and accuracy. The reason is the visibly high number of false positives and false negatives in the shape-only and feature-only models.

The long version of the test data as shown in Figure 13(a) is a 24-hour one-dimensional time series. Figure 13(b) shows a visual summary of the classification results.

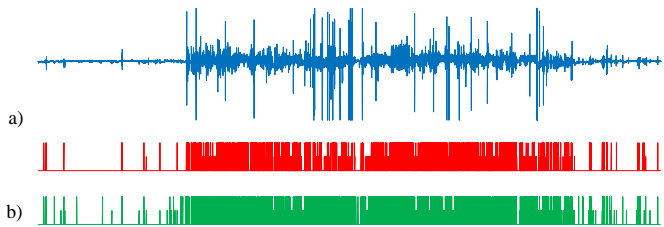

Figure 13: Twenty-four hours chicken test data (*Blue*) annotated with our algorithm (*Green*).

We did not have ground truth annotations for this dataset. The reference labels only show the estimated regions for the behaviors. Moreover, it is difficult, even for an expert, to define precisely where a behavior begins and ends. Nevertheless, it is clear that our algorithm was able to classify most instances of the behaviors in the dataset. As shown in Figure 13(b), our labels (*Green*) and the reference labels (*Red*) are in strong agreement.

Another intuitive way to validate our results is to show that their distribution matches a normal chicken's behaviors. Most animals have a daily recurrent pattern of activity called a "Circadian Rhythm". We examined the existence of such pattern in our results. Figure 14 shows the frequency of each behavior in our results over the course of 24 hours, computed with a 15 minutes long sliding window.

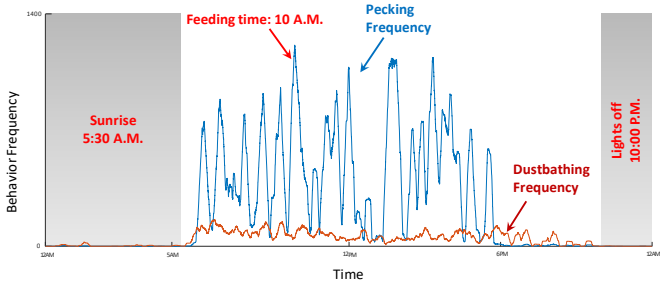

Figure 14: Frequency of chicken behaviors for the 24 hours chicken dataset.

As expected, the `pecking` behavior has the highest overall frequency and it peaks at 10 A.M., which is the feeding time. Also, there is also almost no activities before sunrise and after the artificial light goes off. The pecking behavior is very "bursty". This may be slightly unintuitive but is a familiar fact for anyone with experience with poultry farming.

**5.4 On the Expressiveness of Our Model.** It is fair to say that our proposed method is expressive since the only difference between our algorithm and the other two methods (i.e. shape-based classification and feature-based classification) is the way we combined those two possibilities. Nothing else has changed. Therefore, we can attribute any success only to the increased expressiveness of our model. It might happen that if we put our model in another classification method such as decision tree, it wouldn't work as well for a certain dataset.

## 6 Conclusions and Future Work

We have shown that classifying time series using both shape and feature measures is useful for some data sets, a fact that seems underappreciated in the community. To our knowledge, all relevant works in the literature have adopted either shape-based classification or the feature-based classification approach. We have described a method to create models for different classes in a dataset based on a combination of shape and feature and tested our proposed algorithm on real datasets from different domains. We showed that our method offers significant improvements.